\documentclass{article}

\usepackage{PRIMEarxiv}

\usepackage[utf8]{inputenc} 
\usepackage[T1]{fontenc}    
\usepackage{hyperref}       
\usepackage{url}            
\usepackage{booktabs}       
\usepackage{amsfonts}       
\usepackage{nicefrac}       
\usepackage{microtype}      
\usepackage{lipsum}
\usepackage{fancyhdr}       
\usepackage{graphicx}       
\graphicspath{{img/}}     

\usepackage{caption}
\usepackage{subcaption}
\usepackage{amsmath}
\usepackage{multicol}

\title{Safety Enhancement in Planetary Rovers: \\
Early Detection of Tip-over Risks Using Autoencoders
}

\author{
  Mariela De Lucas Alvarez \\
  Robotics Innovation Center \\
  German Research Center for Artificial Intelligence \\
  Bremen\\
  \texttt{mariela.delucas@dfkide} \\
}

\begin{document}
\maketitle

\begin{abstract}
Autonomous robots consistently encounter unforeseen dangerous situations during exploration missions. The characteristic rimless wheels in the AsguardIV rover allow it to overcome challenging terrains. However, steep slopes or difficult maneuvers can cause the rover to tip over and threaten the completion of a mission. This work focuses on identifying early signs or initial stages for potential tip-over events to predict and detect these critical moments before they fully occur, possibly preventing accidents and enhancing the safety and stability of the rover during its exploration mission. Inertial Measurement Units (IMU) readings are used to develop compact, robust, and efficient Autoencoders that combine the power of sequence processing of Long Short-Term Memory Networks (LSTM). By leveraging LSTM-based Autoencoders, this work contributes predictive capabilities for detecting tip-over risks and developing safety measures for more reliable exploration missions. 

\end{abstract}


\section{Introduction}

\begin{figure}[ht!]
	\centering
	\includegraphics[width=.6\textwidth]{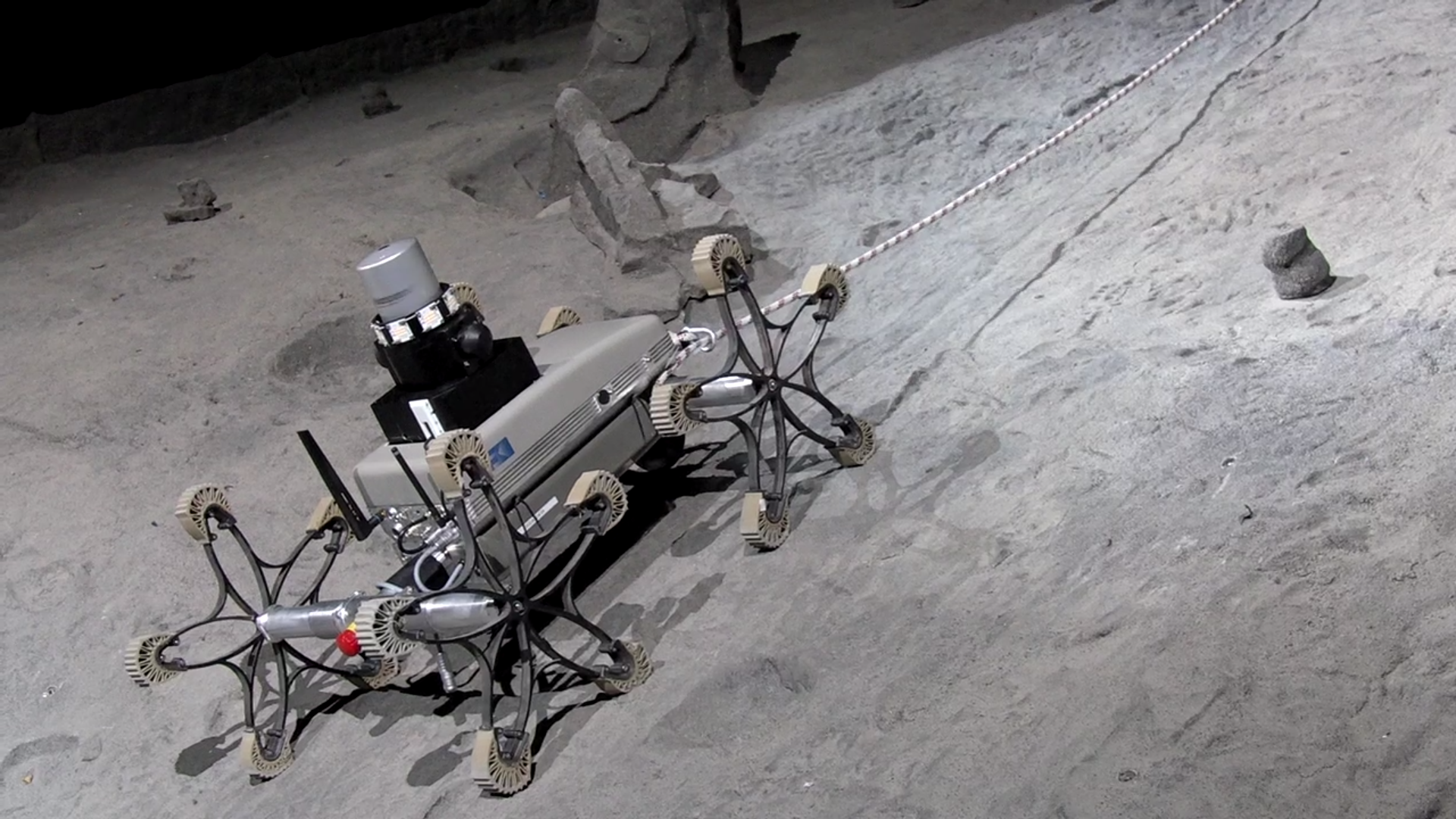}
	\caption[AsguardIV.]{The AsguardIV rover shown traversing an inclined lunar analog plane during data collection.}
	\label{fig:asguard}
\end{figure}

Autonomous exploration robots consistently encounter unexpected events during mission execution. Anticipating such events is crucial for triggering preventive actions to minimize damage to the rover or ensure mission completion. While an exhaustive compendium of the possible dangers an autonomous robot might face is unfeasible to validate, it is crucial to equip the system with methods that detect such events for timely measures to ensure the integrity of the rover and mission.
The AsguardIV is a planetary exploration rover that traverses challenging terrain using rimless wheels that can overcome obstacles (\cite{Dominguez2018,Machowinski2017,HidalgoCarrio2016EnviReE}). The AsguardIV rover is equipped with four rimless wheels, designed to help it navigate challenging terrains where traditional round wheels would slip too much (Figure \ref{fig:asguard}). These uniquely shaped wheels create an undulating motion, even on flat surfaces. This design is intended to facilitate navigation in unstructured environments, such as craters or caves on other planets or moons. Planetary exploration rovers typically navigate conservatively, however, steep slopes and challenging maneuvers during the exploration can lead to tip-over events.

This work explores how such tip-over events can be anticipated with anomaly detection using autoencoders for timeseries forecasting. Autoencoders are particularly effective for anomaly detection due to their ability to learn compact and meaningful representations of data. By training on normal operating conditions of the rover, the Autoencoders in this work can accurately reconstruct expected outputs and highlighting deviations. This makes them ideal for identifying anomalies in sequences, where subtle changes over time can indicate critical issues. Their ability to process and learn from sequential data allows autoencoders to detect unusual patterns and outliers that may signal potential faults or irregularities in complex systems. By learning normal orientation data coming from an Inertial Measurement Unit (IMU), expected orientations can be forecast, and any abnormal patterns that deviate from the learned normal behaviors can be captured.
An approach that combines Long Short-Term Memory networks (LSTM) \cite{Hochreiteretal1997} and Autoencoder frameworks is presented.  The Autoencoder consists of two parametric functions, an encoder and a decoder, that can be trained to reduce reconstruction loss, allowing it to work with any neural network architecture. The Autoencoder model serves as a framework for using LSTMs to create a sequence-to-sequence generator to forecast IMU sequences. Following this, an anomaly detector applies thresholds to these sequences to identify imminent tip-over events. This problem gives rise to two primary objectives:

\begin{enumerate}
\item How far ahead can the IMU be forecast?
\item How accurately can a tip-over event be anticipated?
\end{enumerate}

The first question refers to the maximum time interval into the future for which the IMU data can be accurately predicted, referring to the duration for which reliable forecasts of the IMU readings can be made. This indicates the predictive horizon of the model which is crucial for taking preventive actions.
The second question deals fundamentally with the degree of precision and reliability with which it can be predicted when a tip-over event is about to occur. This concerns the effectiveness and accuracy of the Autoencoders and detection mechanisms used to forecast a sequence and detect anomalies.
Several models are defined to achieve these objectives, thereby establishing new benchmarks in tip-over detection for planetary rovers and enabling the observation of tip-over events from their early stages.

\section{Related Work}
\label{sec:rel_work}

The topic of anomaly detection for space exploration is well-motivated due to its high-stakes applications. Numerous problems have been explored in this area, such as tip-over and slip detection, as well as sensor fault monitoring. The latter can encompass detection, isolation, anticipation and mitigation of intrinsic sensor failures or deteriorating performance. While this is not the primary focus of this work, some of the techniques used support some aspects of the work presented here.

Tip-over detection that considers mobile manipulator stabilization based on Moment Height stability, which accounts for dynamical influences, has been explored in the work of \cite{Fuentes2023_tipover}. Similarly, the work of \cite{Li2021} and \cite{Ding2019} investigates the manipulation of dual-arm robots, focusing on model-based approaches for tip-over analysis and avoidance. These studies include the online evaluation of dynamic stability, considering interactions between the wheel-terrain and vehicle-manipulator systems. Additionally, \cite{Ghosal2012} conducted tip-over stability analysis by performing a series of experiments on a rover with wheel lateral tilt capabilities. The authors measured force-angle stability to assess the rover's ability to traverse uneven terrains safely.

Some interesting works in the field of risk classification are relevant to mention. In the work by \cite{Bouguelia2017} the authors introduce an unsupervised method for classifying by clustering of slip events in a rover based on proprioceptive sensors. The authors use a traditional method, Bayesian tracking, for updating and improving the parameters of the models as new input data comes in. The authors report their method outperforming a K-means solution with a score of $86\%$ vs $80\%$ accuracy. The success of this work establishes that it is possible to design and implement unsupervised models to detect unforeseen and unstructured events.

The following works introduce the importance of monitoring traction in planetary exploration rovers as it can negatively impact the mission by immobilizing a vehicle. The work by \cite{Gonzalez2018a} addresses the issue of slippage. Their approach consists of evaluating two supervised machine learning methods, SVM and Neural Network, against two unsupervised learning approaches, K-Means and Self-Organizing Maps (SOM). Their experiments were performed on a single-wheel test-bed with IMU sensors placed where the detection would be maximized in the chassis. Results show that a SOM-based algorithm balances the advantages of supervised and unsupervised learning algorithms which are high success rate and low storage requirements respectively.

The authors later extend their work in \cite{Gonzalez2019} by performing rigorous tuning to evaluate their algorithms with 55 configurations. They also incorporate other types of settings that they consider can influence the performance of the models. Such are rover speed, type of terrain and type of tire. In a different publication (\cite{Kruger2019}) they increase settings evaluation like a sandy incline, more slip classes, different rover velocities and sensor inputs. Their SOM solution still outperforms compared to other algorithms.

Some works focus on the importance of slip detection \cite{Angelova2007,Helmick2009} by treating it as an important metric for mobility risk prevention. An interesting work by \cite{Skonieczny2019} recognizes the issue of mobility risks as one of the strongest and most used relationships in mobility prediction. This work focuses on explicitly assessing mobility risk vs slope by defining a metric that uses high slip fraction. This is defined as the proportion of slip data points above a certain percentage mark. Additionally, this work also performs terrain assessment on different types of terrain. This combination of methods helps in the early detection of potential risks and a good balance between rover responsiveness and stability.

Some research considers the problem of terrain identification for survey rovers as a basis for risk identification. In \cite{Banerjee2020} the authors developed a framework that models wheel-terrain dynamics in a rover that is able to adapt rapidly to the change of terrain. The authors have done so by capturing non-linear interactions in the dynamics of the controller and the terrain. The linear model is augmented with new data learned with features learned from a Neural Netowrk (NN) with a Bayesian regression algorithm.

The same problem is approached in \cite{Dimastrogiovanni2020} where a comprehensive set of proprioceptive sensors are used to classify terrain. Mobility risks are subsumed under the category of slippage and are treated as a type of terrain that can be detected by the classifier. The authors implement an SVM using on their own collected data, reporting robust results.

These works addressed tailored needs concerning anomaly perception. Methods applied to space rovers specifically do acknowledge the relevancy of terrain uncertainty and therefore utilize different methods to prevent events that jeopardize missions and equipment. 
This motivates the interest in leveraging the power of LSTMs to detect potential imminent risk events without requiring any prior knowledge of the terrain or vehicle dynamics. 

\section{Methodology}
\label{sec:methods}

Autoencoders are a type of NN architecture composed of two signal-processing components: an Encoder and a Decoder. The Encoder encodes an input signal into a latent or different space, while the Decoder takes this latent representation generated by the encoder and recreates the original input signal.

\begin{figure}[t]
    \centering
    \includegraphics[width=.9\textwidth]{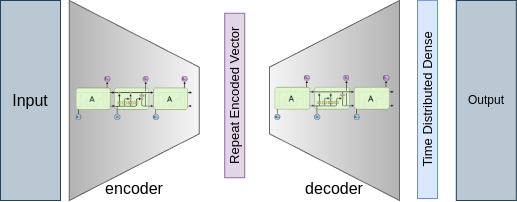}
    \caption[Architecture of the LSTM Autoencoder.]{Architecture of the LSTM Autoencoder. Using LSTM layers enables the Autoencoder to learn temporal structures to forecast sequences from reconstructed inputs \cite{mythesis}.}
    \label{fig:enc_dec}
\end{figure}

This framework has been most pervasively used as sequence-to-sequence solutions in various applications most notably in earlier applications of Natural Language Processing (\cite{SutskeverVL14,Cho2014}). Autoencoders architectures can be designed using any neural network to compress and recreate data. Figure \ref{fig:enc_dec} illustrates how LSTMs are used as Autoencoders in this work to compress time-distributed latent representations. The use of LSTMs is advantageous as they overcome the vanishing-exploding gradient problem \cite{Goodfellow-et-al-2016, Graves:series/sci/2012-385}. This problem arises when the propagation of the gradients through stack layers causes the gradient to become imperceptible or to inflate excessively, making it extremely difficult to adjust the weights appropriately and causing the network to stop learning. This instability can impact the learning of sequential dependencies.
Using LSTMs in this application helps mitigate this problem as the Autoencoder needs to be able to accurately encode and regenerate the input data for sequence forecasting to perform anomaly detection. 

The learning challenges addressed by this dataset involve forecasting and anomaly detection.  While LSTM's are effective for timeseries forecasting, some considerations must be made beforehand. A LSTM alone can model one-step forecasting problems, where only the next time step is predicted, but for multi-step forecasts, an Autoencoder is preferred. The reason is because in a LSTM network, the number of nodes in the output layer must match the number of output time steps, which isn't optimal for describing specific time steps for a sequence. This problem requires multi-step forecasts, as a tip-over risk spans multiple time steps. The Autoencoder allows for a more interpretable output , accommodating the multi-step forecast requirement more effectively.

\paragraph{Sequence Composition}
Sequence forecasting is what is fundamentally a sequence-to-sequence problem. With an Autoencoder, it is possible to address sequence forecasting by harnessing the time series processing capabilities of the LSTM. The LSTM layers learn the relationship between the time steps in the input as encoded sequences. The decoder then uses the encoded sequences to generate the output. In this case the length of the input and output sequences are different, which the Autoencoder framework is capable of handling.

The sensor readings are transformed into a format suitable for the forecasting problem. The requirement is that for a given past sequence of readings $X^{i}_{t-n}, ..., X^{i}_{t-1}$ the current $n$ time steps $X^{i}_{t}, ..., X^{i}_{t +n}$ is forecast. This is framed into a sequence-to-sequence problem, where the first sequence is given as input into a model that learns to predict the subsequent sequence as output. This applies to all time steps with the superscript $i$ as the axis ID of each sensor. 

\paragraph{Data}
The data used in this work is a collection of survey missions specifically gathered for tip-over risk detection with the AsguardIV rover. This rover executed both autonomous and manual navigation in a moon crater analog environment. The analog was constructed based on real moon images taken by Apollo missions and includes challenges such as slopes with varying inclinations, rocks, and boulders. The rover traversed surfaces with inclinations ranging from 15$^{\circ}$ to 35$^{\circ}$.

The dataset comprises sensor readings from 10 trials, totaling approximately 250,000 samples. These readings come from the rover's calibrated sensor module, which transmits acceleration and orientation data along the three axes (x, y, z) of the IMU at approximately 100Hz. The data, originally in log format, contained various types of information beyond the sensor readings, necessitating the extraction of messages that contained only the navigational data. The most relevant data for identifying the goal behavior in this case is the IMU data. This includes calibrated sensor readings from the accelerometer and gyroscope in three axes, conforming to a multivariate vector of six features:

The input-output sequences are a multivariate vector of six features:
\begin{equation}
X_{t} = \{accel^{x}_{t}, accel^{y}_{t}, accel^{z}_{t}, gyro^{x}_{t}, gyro^{y}_{t}, gyro^{z}_{t} \}
\end{equation}

The main objective of these recordings was to capture normal operation and tip-over risk scenarios as the rover traversed uneven terrain. The data is unlabeled for training since it is an unsupervised learning setting. However, each log is annotated with the observable mobility conditions it recorded. The logs contain the following behaviors:

\begin{enumerate}
    \item \textit{Normal.} No mobility risks during navigation.
    \item \textit{Tip-over risk.} When the rover begins to tip-over.
    \item \textit{Slip.} When the rover slides while navigating.
\end{enumerate}

\begin{figure}[t]
    \centering
    \captionsetup{type=figure}
    \begin{subfigure}{0.35\textwidth}
        \centering
        \includegraphics[width=\textwidth]{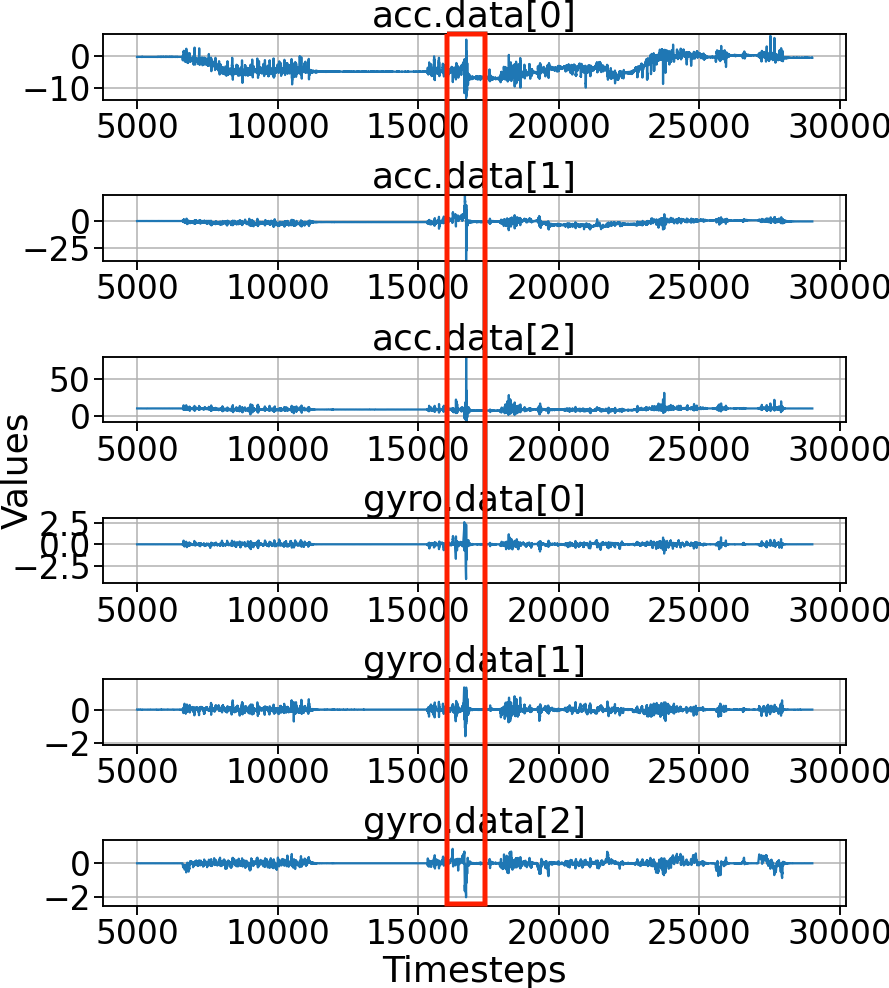}
        \caption{Tip-over Risk}
        \label{fig:hazard}        
    \end{subfigure}
    \begin{subfigure}{0.35\textwidth}
        \centering
        \includegraphics[width=\textwidth]{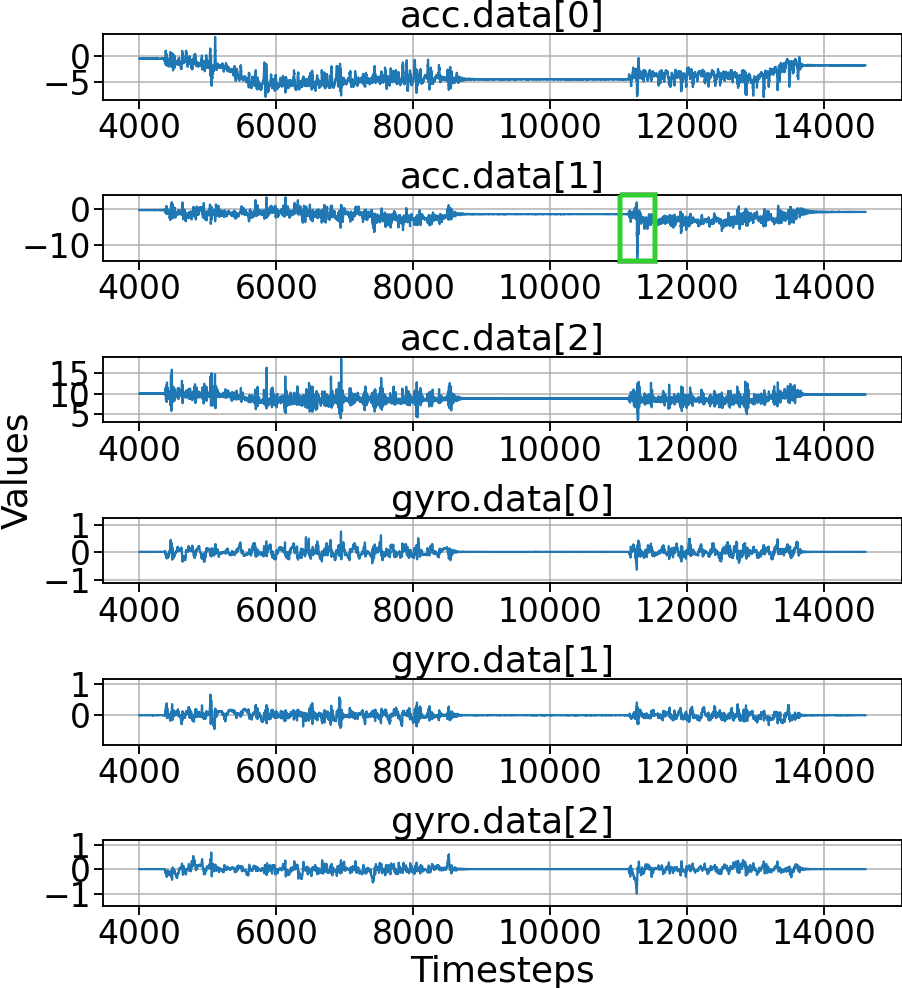}
        \caption{No Risk}
        \label{fig:no_hazard}
    \end{subfigure}
    \caption[Two AsguardIV Trials.]{Examples of the AsguardIV performing two navigation executions \cite{mythesis}. IMU plots across all axes. Figure (a) illustrates a mobility risk marked in red. Figure (b) shows a navigation sequence with no mobility risks, however, it marks in green a sequence that could be mistaken as one.}
    \label{fig:trials}
\end{figure}

It is not easy to characterize all conditions of the signal patterns of the rover with visual inspection. Figure \ref{fig:trials} shows how these conditions could be confused. The first shows a tip-over incipience that required a manual stop to avoid damage to the rover. The second could be mistaken as mobility risk, however, there was no actual mobility danger, but the rover descended too fast for a brief moment.

\paragraph{Hyperparametrization}
The Autoencoders were tuned with hyper-parametrization method Bayesian Optimization Hyperband (BOHB) (\cite{Falkner2018}), combining Bayesian optimization (\cite{Shahriarietal2016}) and Hyperband (\cite{Li2018}). It has been shown that BOHB optimizes progressively faster than Random Search, Bayesian optimization and Hyperband alone.

Bayesian optimization models an objective function as, $p(f|D)$ based on observed data, $D = {(x_{0}, y_{0}),...,(x_{i-1}, y_{i-1})}$, using an acquisition function $a = X \rightarrow \mathbb{R}$ that balances exploration and exploitation. Hyperband is a hyper-parameter optimization strategy that maximizes budget efficiency by repeatedly utilizing Successive Halving (SH) (\cite{Jamieson2016}) to identify the best configuration out of \textit{n} randomly sampled configurations. SH evaluates the chosen configurations on a given budget and ranks them based on performance. It then continues evaluating the top $\eta^{-1}$ configurations on an $\eta$ times larger budget until the maximum budget is reached.

Hyperband has demonstrated superior performance compared to both Random Search and Bayesian optimization. However, its effectiveness is limited when converging to the global optimum due to the randomly chosen configurations, even when large budgets are chosen. To address this, BOHB integrates the Tree-Parzen Estimator (TPE) (\cite{Bergstra2011}), which is simpler than standard Gaussian Process-Bayesian optimization because the latter often requires complex approximations. Kernel Density Estimators (KDE) are used to model the densities over the input configuration space instead of modeling the objective function $f$ directly by $p(f|D)$. The result is a variety of densities over the configuration space $X$ using different observations ${x^{(1)}, ..., x(k)}$ in the non-parametric densities. TPE defines $p(x|y)$ with two densities,

\begin{equation}
\label{eq:kde_densities}
	p(x|y) = 
		\begin{cases}
			l(x) & \text{if } y<y^{*} \\
			g(x) & \text{if } y \geq y^{*}
		\end{cases}
\end{equation}

In BOHB, $l(x)$ is the density from observations where the loss $f(x^{(i)})$ is less than $y^{*}$, and $g(x)$ is the density from the remaining observations. By not using all observations, BOHB enhances computational efficiency for Hyperband, enabling many function evaluations with small budgets. Unlike TPE, BOHB uses a single multidimensional KDE for density estimation to better handle interaction effects in the input space. 

To calculate the KDE, a minimum of $N_{min}$ data points are required, satisfying $d+1$, where $d$ is the number of hyperparameters. Instead of waiting for $N_{b} = |D_{b}|, N_{min} + 2$ configurations are initialized, and the best and worst configurations model the two densities:

\begin{equation}
\begin{aligned}
\label{eq:quick_densities}
N_{b,l} = \max(N_{min}, q \cdot N_{b})\\
N_{b,g} = \max(N_{min}, N_{b} - N_{b,l})\\
\end{aligned}
\end{equation}

Eexpected Improvement (EI) \cite{Jones2001} is finally optimized by sampling $N_{s}$ points from $l'(x)$, an approximation KDE of $l(x)$ with all bandwidths multiplied by a bandwidth factor of $b_{w}$ to explore promising configurations. The improved convergence and time efficiency motivated the evaluation of many Autoencoder configurations.

\section{Autoencoder Selection}
\label{sec:model_selection}

 \begin{table}[t]
 \begin{minipage}{.4\linewidth}
    \centering
    \caption{General configuration parameters.}
    \label{tab:bohb-config}
    \resizebox{\linewidth}{!}{
    \begin{tabular}{lcc}
        \toprule
        \textbf{BOHB parameter} & \textbf{Values} & \textbf{Unit}\\
        \midrule
        Nodes & $\{4, 64\}$ & - \\
        Maximum budget & 100 & Epochs\\
        \toprule
        \textbf{Sets Configuration} & \textbf{Values} & \textbf{Unit}\\
        \midrule
        Depth & $\{1, 2, 3\}$ & Layers\\
        IO Sequence Length & $\{25, 50, 100\}$ & Timesteps\\
        Step size & $\{None, 10\}$ & Timesteps\\
        Dropout & .20 & $\%$\\
        \toprule
        \textbf{Fixed parameters} & \textbf{Setting} & \\
        \midrule
        Loss & MSE & \\
        Batches & 16 & \\
        Optimizer & Adam& \\
        \bottomrule
    \end{tabular}
    }
\end{minipage}
\hfill
\begin{minipage}{.55\linewidth}
        \centering
        \includegraphics[width=\linewidth]{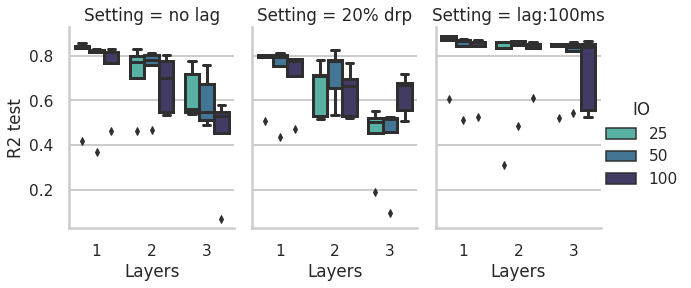}
        \caption{Test performance}
        \label{fig:best_r2_test}
        \captionof{figure}[Best Test $R^{2}$ score.]{First test results showing $R^{2}$ scores per different Autoencoder layer depths per non-BOHB setting.}
        \label{fig:best_bohb_test}
\end{minipage}
\hfill
\end{table}

The configuration space for BOHB sampling specifies the number of nodes per LSTM layer, ranging from a minimum of 4 units to a maximum of 64 units. Additional experimental settings outside of BOHB include network depth, sequence length, sequence step, and dropout, as detailed in Table \ref{tab:bohb-config}.

Three sets of optimization experiments are generated, characterized by a sampling step, no step, and dropout. These are conducted for each input-output sequence length per layer depth, totaling 27 BOHB runs. The input-output sequence lengths of 100, 50, and 25 correspond to 1 second, and 500 and 250 milliseconds at 100Hz, respectively.

Certain settings remained consistent across all experiments: the loss metric was Mean Squared Erro (MSE), the \textit{ADAM} optimizer was used, and training budgets ranged between 20 and 100 epochs. SH was configured to perform 3 iterations for budget allocation. A total of 216 unique configurations were evaluated, with 135 using the full budget. Initial results, grouped by non-hyperparameter settings, assessed network performance using the $R^{2}$ score, with validation and test scores plotted in Figure \ref{fig:best_bohb_test}.

A general observation is that networks using a stepped sampling sequence format of 100ms exhibit less variance across all candidate sets compared to other settings. Networks with more than one layer and without a stepped input show more variance and outlier scores, likely due to non-tip-over risk behaviors such as slip not being considered normal. A 10ms step achieves more stable scores, particularly in smaller network depths, and consistently outperforms other settings in tests at any network depth.

\begin{table}[t]
    \begin{minipage}{.4\linewidth}
    \centering
    \caption[2nd Experiment.]{Configuration settings in 2nd test for networks using sampling step achieving $R^{2} >=.9 $.}
    \label{tab:experiments_2}
    \resizebox{\linewidth}{!}{
    \centering
        \begin{tabular}{cccc}
            \toprule
            IO (ts) & Layers & Nodes & Test $R^{2}$\\
            \toprule
            25  & 1 & [6] 	   &  0.925\\
            25  & 2 & [40, 12] & 0.929\\
            25  & 3 & [18, 5, 5] &  0.923\\
            50  & 1 & [5] & 0.915\\
            50  & 2 & [14, 37] & 0.903\\
            100 & 1 & [15] & 0.909\\
            100 & 2 & [21, 11] & 0.905\\
            \toprule
        \end{tabular}
    }
    \end{minipage}
    \hfill
    \begin{minipage}{.55\linewidth}
        \caption[3rd Experiment.]{Configuration settings in 3rd test for networks using sampling step achieving $R^{2} >=.9 $.}
        \label{tab:experiments_3}
        \centering
        \resizebox{\textwidth}{!}{
            \begin{tabular}{cccccc}
                \toprule
                Step (ts) & In (ts) & Out (ts) & Layers & Nodes & Test $R^{2}$\\
                \bottomrule
                10  & 25 & 10 & 1 & [4]   & 0.942 \\
                \textbf{10}  & \textbf{25} & \textbf{5} & \textbf{1} & \textbf{[19]} & \textbf{0.969}\\
                none & 25 &  5 & 1 & [48] & 0.937\\
                none & 25 & 10 & 1 & [49] & 0.913\\
                \bottomrule
            \end{tabular}
            }
        
    \end{minipage}
\end{table}

A second set of experiments was performed using $70\%$ of the logs as training data to improve these results of which seven candidates achieved higher than a $.9$ $R^{2}$, summarized in Table \ref{tab:experiments_2}. The common setting for these networks is a 100ms sampling step. There appears to be no improvement in score despite the varying number of cells and layers among the networks. This confirms that a shorter input sequence of 250ms input with a 100ms step consistently performs better than others in any layer depth. In addition, having a single layer network consistently performs better or comparable to deeper networks within the same setting.

This motivated a third set of experiments reducing the length of the output sequence to $5$ time. The same initialization parameters from Table \ref{tab:bohb-config} are used but this time the network depth was set to $1$ layer. The final performance results are shown in Table \ref{tab:experiments_3}.

This evaluation showed a $4\%$ from the first BOHB experiments. These findings help in selecting a potential deployment model, answering the question: \textit{How far ahead can the IMU be forecast?}. Initially, it was speculated that some networks could forecast up to 1 second in advance. However, it was observed that LSTMs perform best with shorter input sequences and a short horizon output. The models showed that the best performance is achieved with a 250ms input signal and a 50ms output signal.


\begin{figure*}[t]
    \centering
    \begin{subfigure}[t]{0.33\textwidth}
        \centering
        \includegraphics[width=\textwidth]{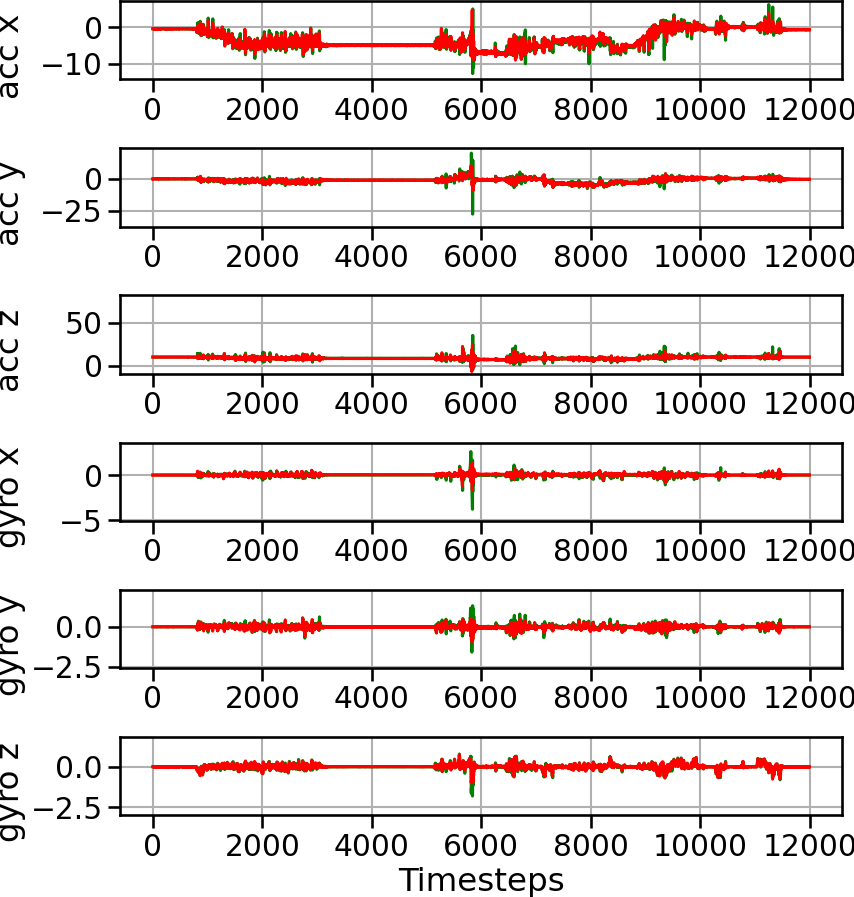}
        \caption{Forecast of test trial 1}
        \label{fig:forecast_test_0}
    \end{subfigure}
    \hfill
    \begin{subfigure}[t]{0.33\textwidth}
        \centering
        \includegraphics[width=\textwidth]{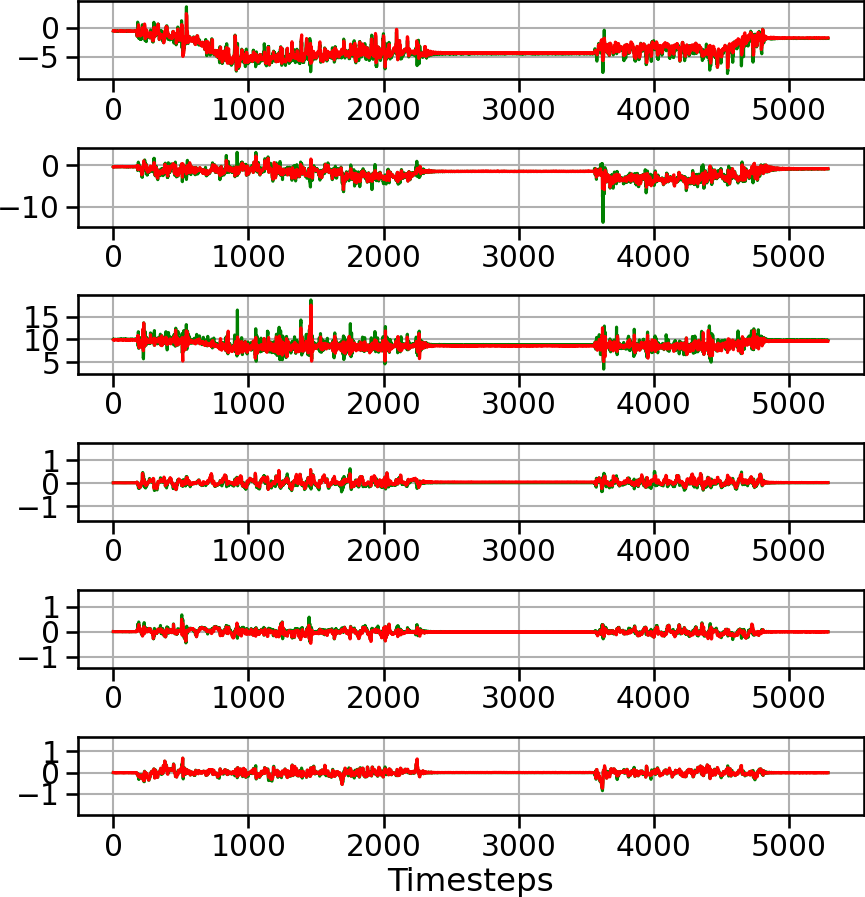}
        \caption{Forecast of test trial 2}
        \label{fig:forecast_test_1}
    \end{subfigure}
    \hfill
    \begin{subfigure}[t]{0.33\textwidth}
        \centering
        \includegraphics[width=\textwidth]{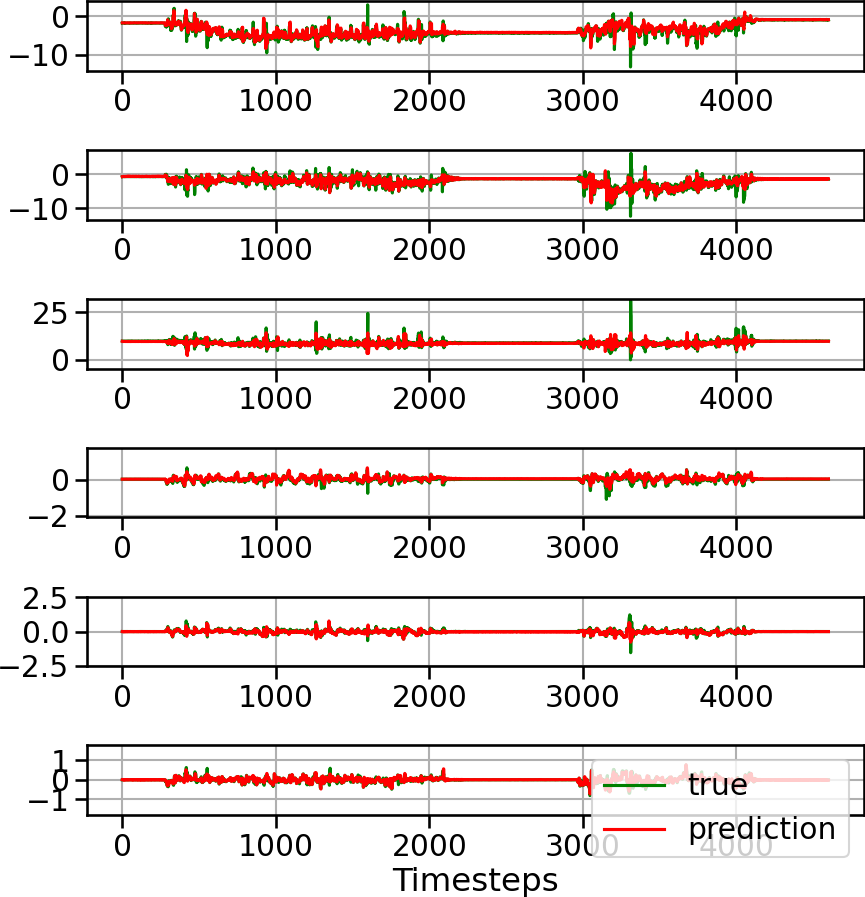}
        \caption{Forecast of test trial 3}
        \label{fig:forecast_test_2}
    \end{subfigure}
    \caption{Tests trials showing the forecast signals (red) from the in put sequence (green). It can be observed that the normal IMU signals are accurately predicted whereas the anomalies are not, enabling effective detection of potential tip-over risks. }
    \label{fig:test_forecasts}
\end{figure*}

\section{Tip-over Risk Detection}

The reliability of the tip-over risk detection is now evaluated based on the robustness of the IMU forecast. Three test trials have been selected for risk conditions detection, these include two tip-over risks and one slip event. The selected Autoencoder can accurately forecast IMU signals as shown in Figure in \ref{fig:test_forecasts} for each of the test trials using the best model with $R^{2}$ score of $0.969$. The progression of each navigation test is unfolded as follows:

\paragraph{First Test Trial.} A goal point was set near the middle of a slope and a waypoint navigation trajectory was generated for autonomous navigation. The first goal point was reached. However, the rover experienced a severe downward motion during descent, almost tipping over. This event is illustrated in Figure 5(a), shortly before timestamp $ts=6,000$. At this critical juncture, the mission was halted, and the rover was manually taken down.

\paragraph{Second Test Trial.} This trajectory was performed in manual navigation mode and did not present any risk events but did include a slipping motion. The mission was planned as a forward ascent followed by a backward descent. During the descent, the rover experienced a slight slip in a crater but completed a smooth backward climb descent. This event can be observed near timestamp $ts=3,500$.

\paragraph{Third Test Trial.} This trial followed a similar plan as the second, starting from a different location. During the descent, one wheel twisted upwards, almost causing the rover to tip over. This near-tip-over event is observed around timestamp $ts=3,250$




\begin{figure}[t]
    \centering
    \begin{subfigure}[t]{0.47\textwidth}
        \centering
        \includegraphics[width=\textwidth]{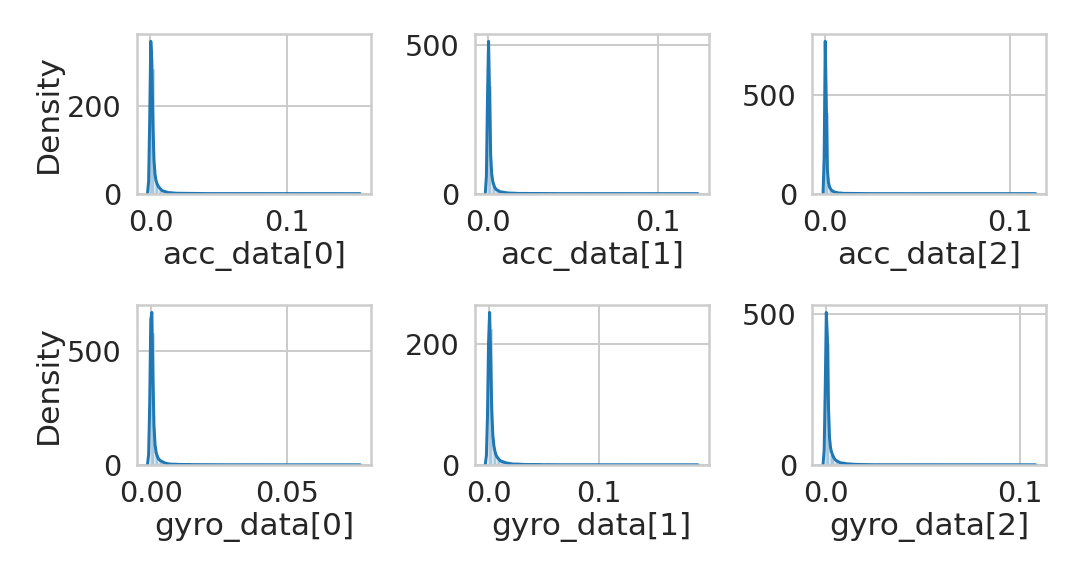}
        \caption{MSE}
        \label{fig:MSEdist_plot}
    \end{subfigure}
    \hfill
    \begin{subfigure}[t]{0.47\textwidth}
        \centering
        \includegraphics[width=\textwidth]{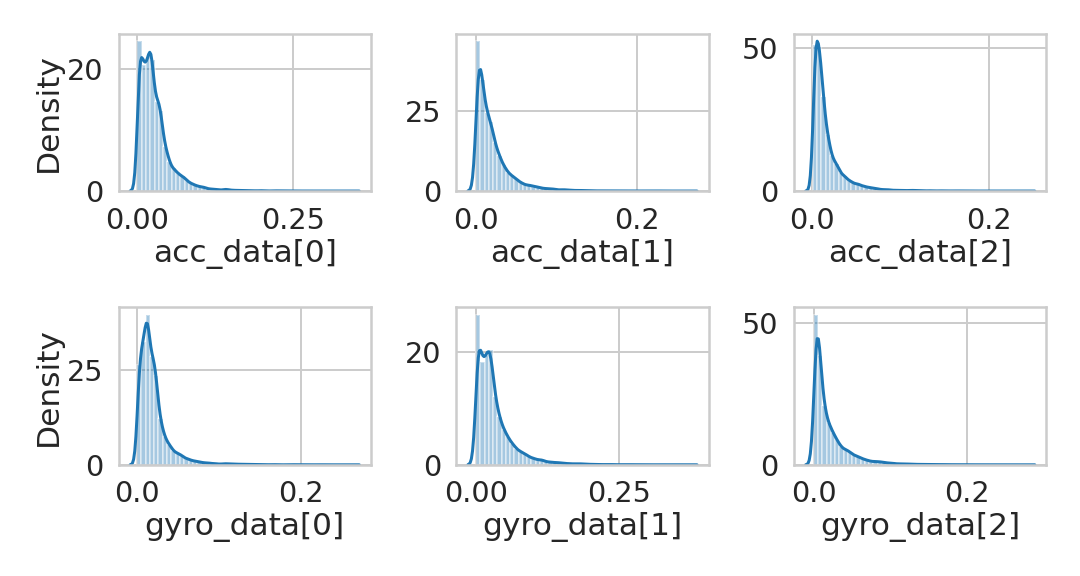}
        \caption{MAE}
        \label{fig:MAEdist_plot}
    \end{subfigure}
    \caption[MAE and MSE loss distributions.]{(a) MSE and (b) MAE IMU predictions distribution for the selected Autoencoder. Each plot shows data for each sensor axis.}
    \label{fig:Loss_Distributions}
\end{figure}

\begin{table}[t]
	\centering
        \caption[Selected loss thresholds.]{Selected thresholds for each loss based on the IMU prediction distributions.}
	\label{tab:loss_thresh}
	\resizebox{.5\linewidth}{!}{
		\begin{tabular}{ccccccc}
			\toprule
			Loss & $accel^{x}$ & $accel^{y}$ & $accel^{z}$ & $gyro^{x}$ & $gyro^{y}$ & $gyro^{z}$\\
			\toprule
			MSE & 0.04 & 0.04 & 0.04 & 0.02 & 0.05 & 0.04 \\
			MAE & 0.25 & 0.25 & 0.25 & 0.25 & 0.25 & 0.25 \\
			\toprule
		\end{tabular}
		}
	
\end{table}

To evaluate tip-over risk detection for the selected Autoencoder, the prediction test errors, Mean Squared Error (MSE) and Mean Absolute Error (MAE), are calculated for the predicted output as shown in Equations \ref{eq:mse} and \ref{eq:mae}:

\begin{equation}
MSE = \frac{1}{n} \sum^{n}_{i=1} (y_{i} - \hat{y_{i}})^{2} \\
\label{eq:mse}
\end{equation}

\begin{equation}
MAE = \frac{1}{n} \sum^{n}_{i=1} |y_{i} - \hat{y_{i}}| \\
\label{eq:mae}
\end{equation}

Fundamentally, anomalies are rare and it is assumed that these can be found at the edges of data distributions. The MAE and MSE distributions of the IMU predictions from the training data are used to establish thresholds for risk detection. Figure \ref{fig:Loss_Distributions} shows these densities for each sensor reading. Based on these distribution plots, a set of thresholds is defined for each sensor for each loss type as shown in Table \ref{tab:loss_thresh}.

\begin{figure}[t]
    \centering
    \begin{subfigure}[t]{0.33\textwidth}
        \centering
        \includegraphics[width=\textwidth]{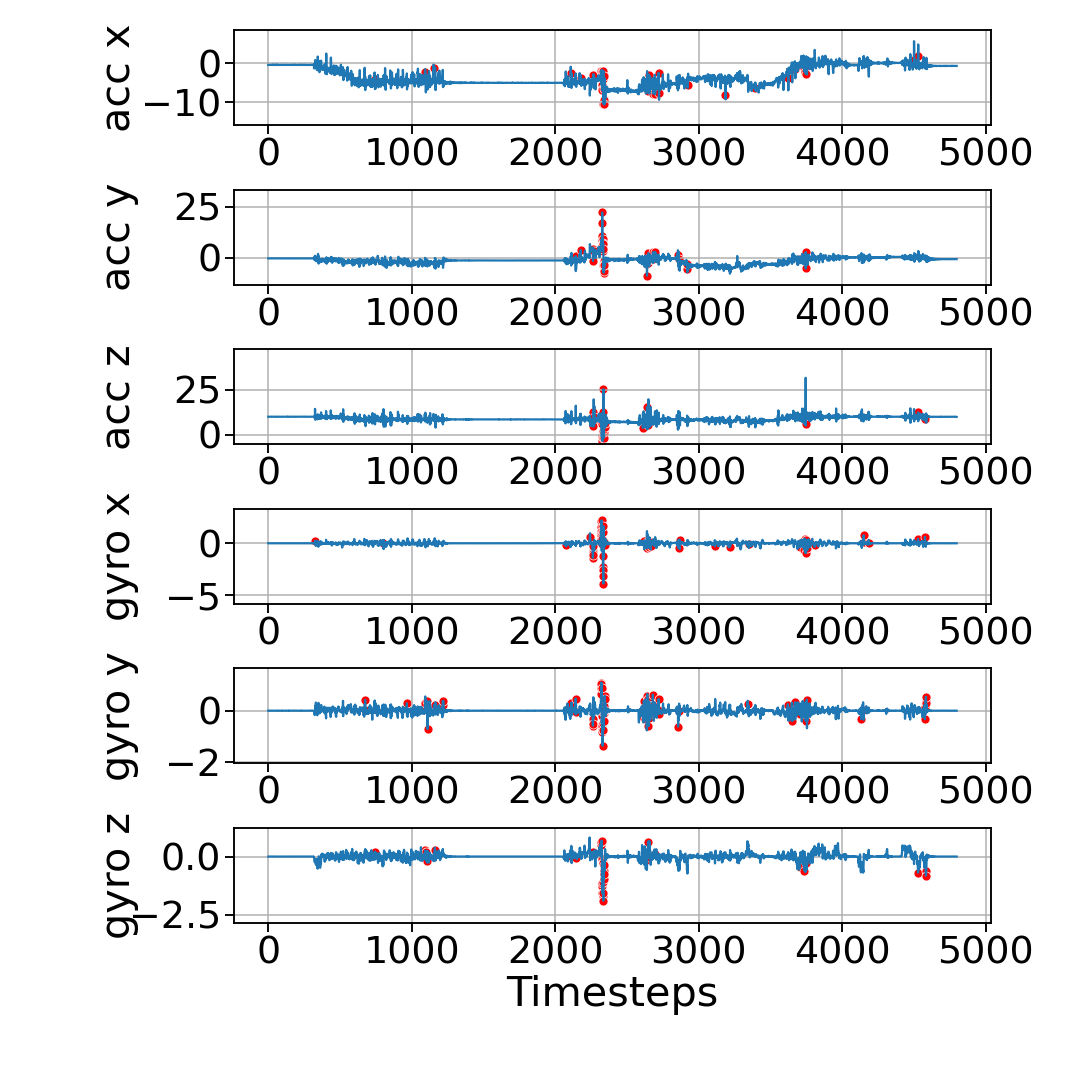}
        \caption{Test trial 1}
        \label{fig:trial0_MSE_anomaly_plot}
    \end{subfigure}
    \hfill
    \begin{subfigure}[t]{0.33\textwidth}
        \centering
        \includegraphics[width=\textwidth]{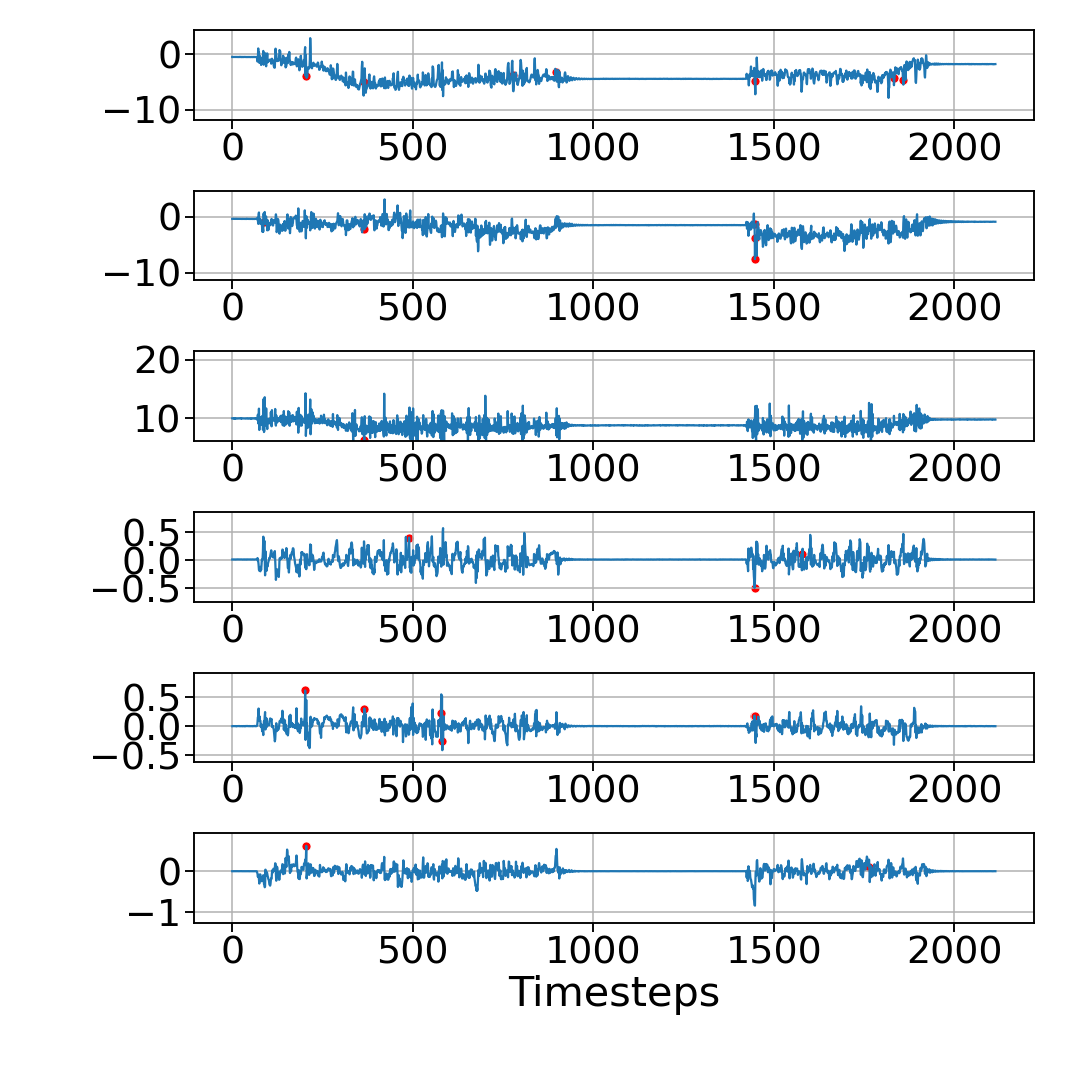}
        \caption{Test trial 2}
        \label{fig:trial1_MSE_anomaly_plot}
    \end{subfigure}
    \hfill
    \begin{subfigure}[t]{0.33\textwidth}
        \centering
        \includegraphics[width=\textwidth]{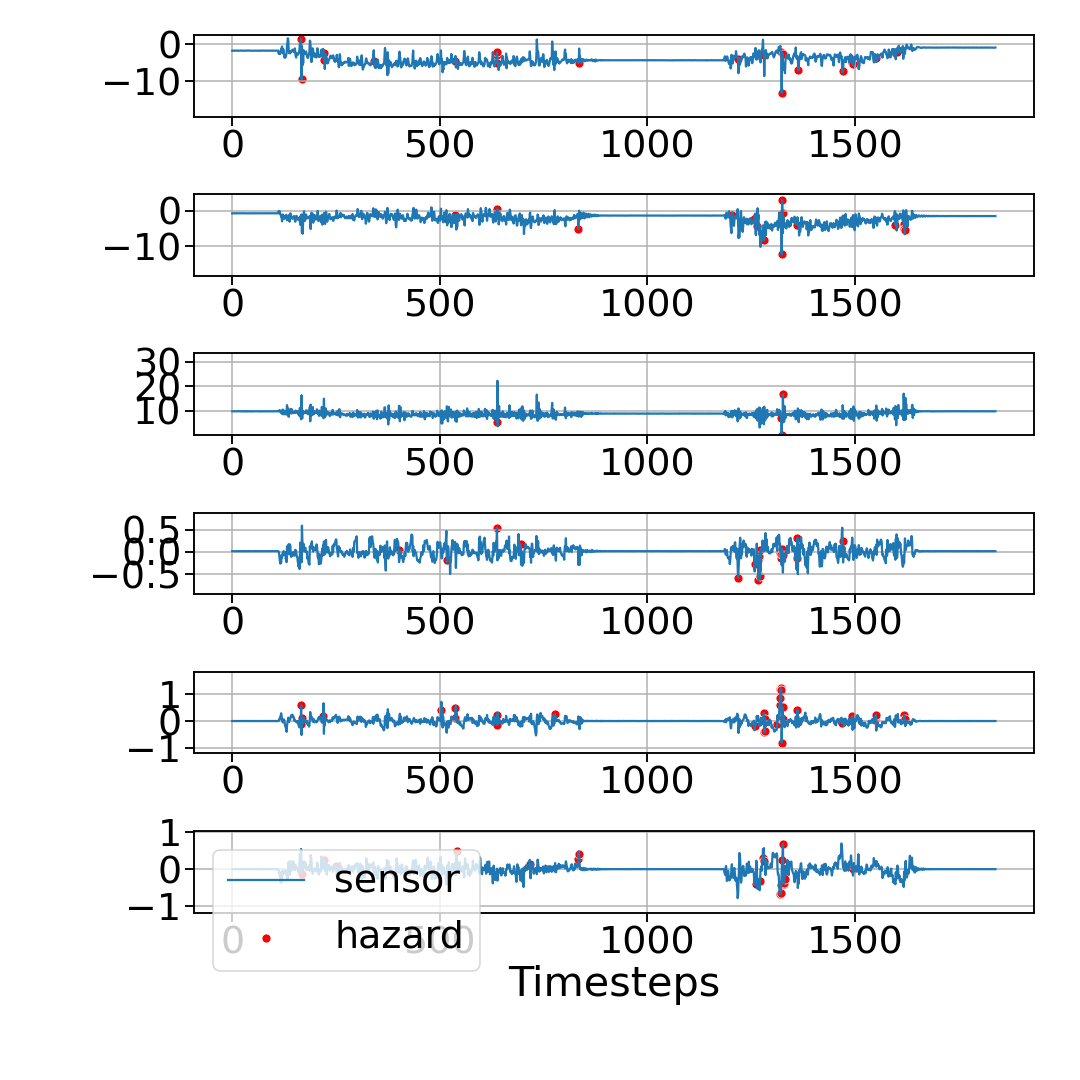}
        \caption{Test trial 3}
        \label{fig:trial2_MSE_anomaly_plot}
    \end{subfigure}
    \caption[Risk Detection with MSE]{Tip-over risk detection using MSE threshold values for three test trials. IMU sequences from three test trials in blue, with red dots marking the time steps where the predictive error exceeded the detection threshold.}
    \label{fig:detection_mse}
\end{figure}

\begin{figure}[t]
    \centering
    \begin{subfigure}[t]{0.33\textwidth}
        \centering
        \includegraphics[width=\textwidth]{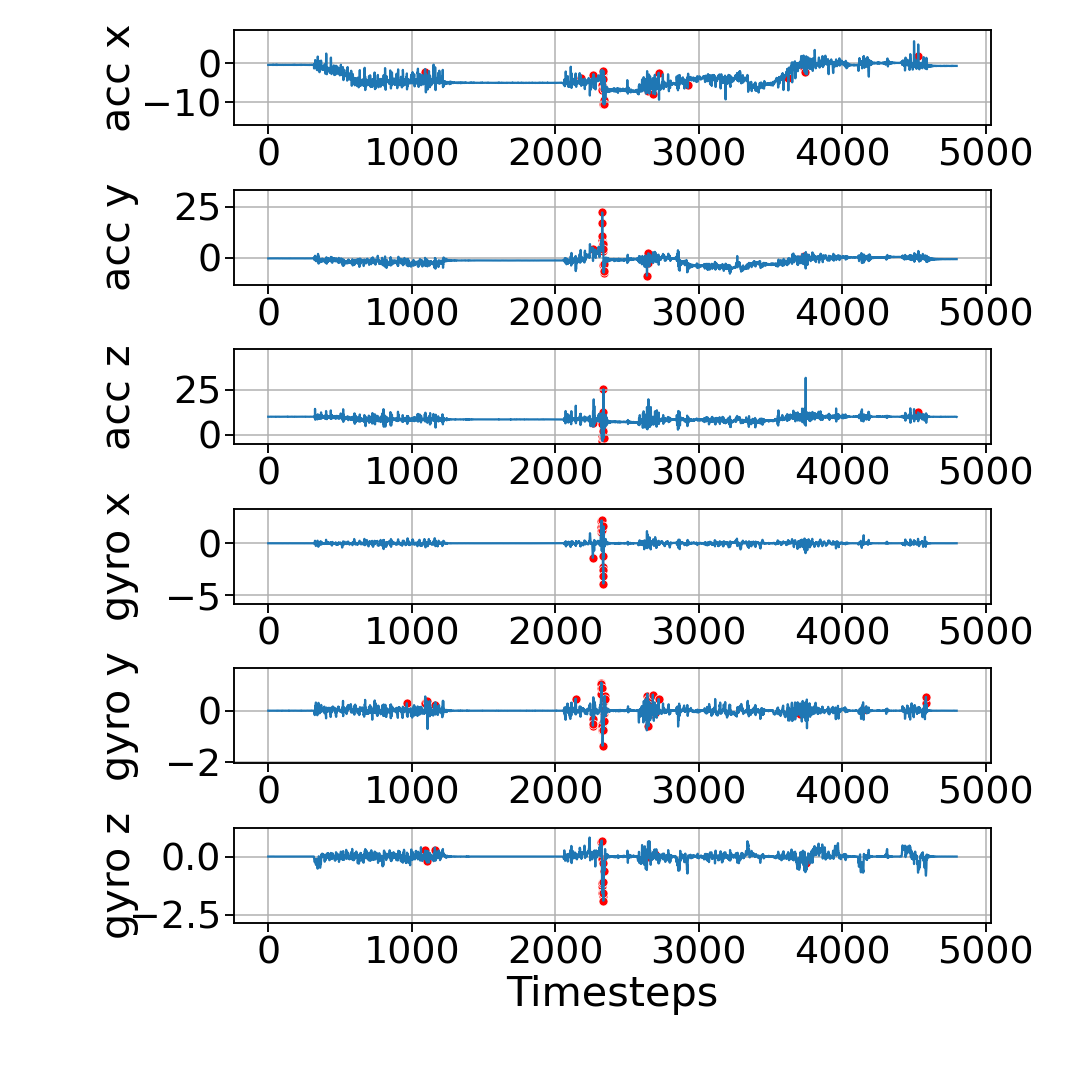}
        \caption{Test trial 1}
        \label{fig:trial0_MAE_anomaly_plot}
    \end{subfigure}
    \begin{subfigure}[t]{0.33\textwidth}
        \centering
        \includegraphics[width=\textwidth]{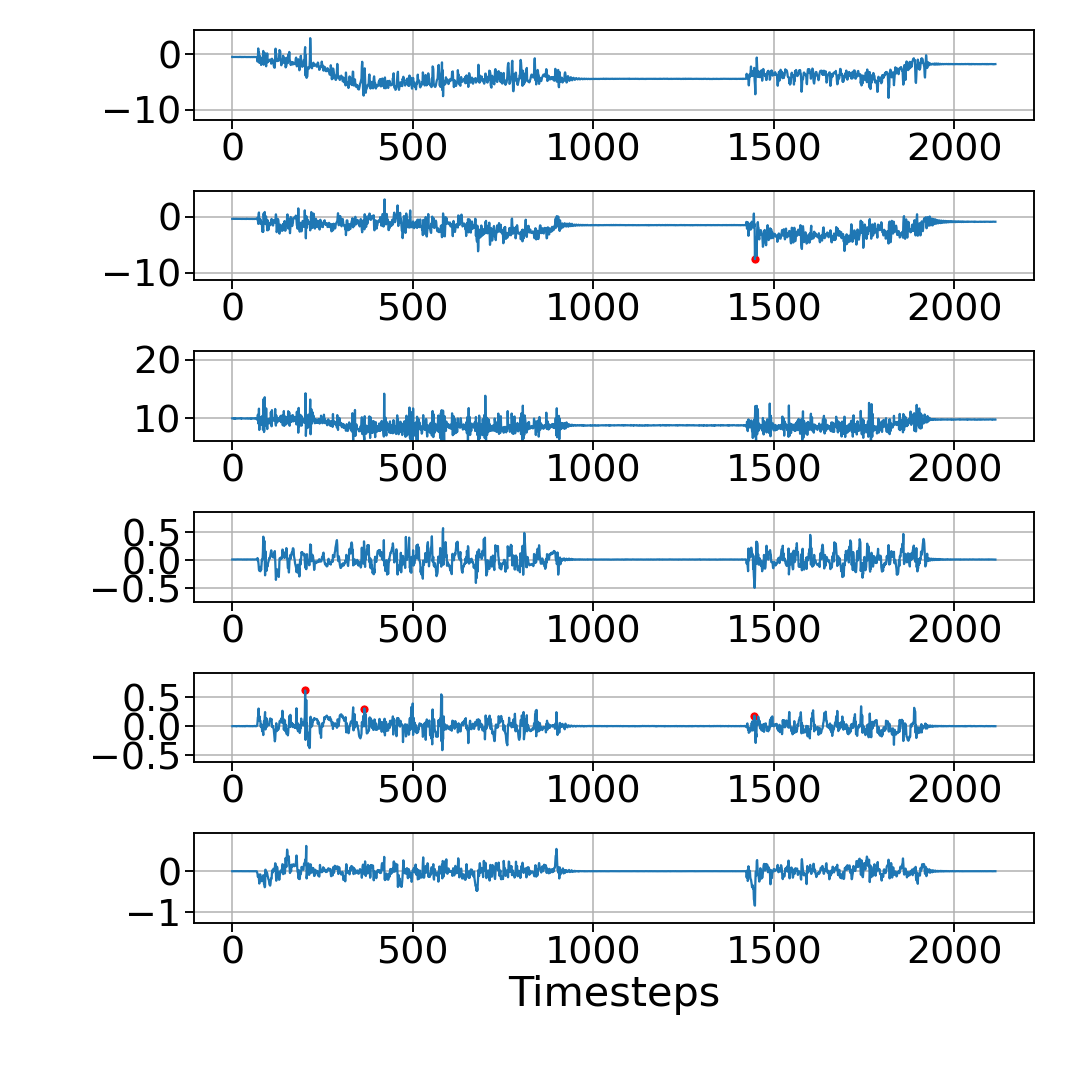}
        \caption{Test trial 2}
        \label{fig:trial1_MAE_anomaly_plot}
    \end{subfigure}
    \begin{subfigure}[t]{0.33\textwidth}
        \centering
        \includegraphics[width=\textwidth]{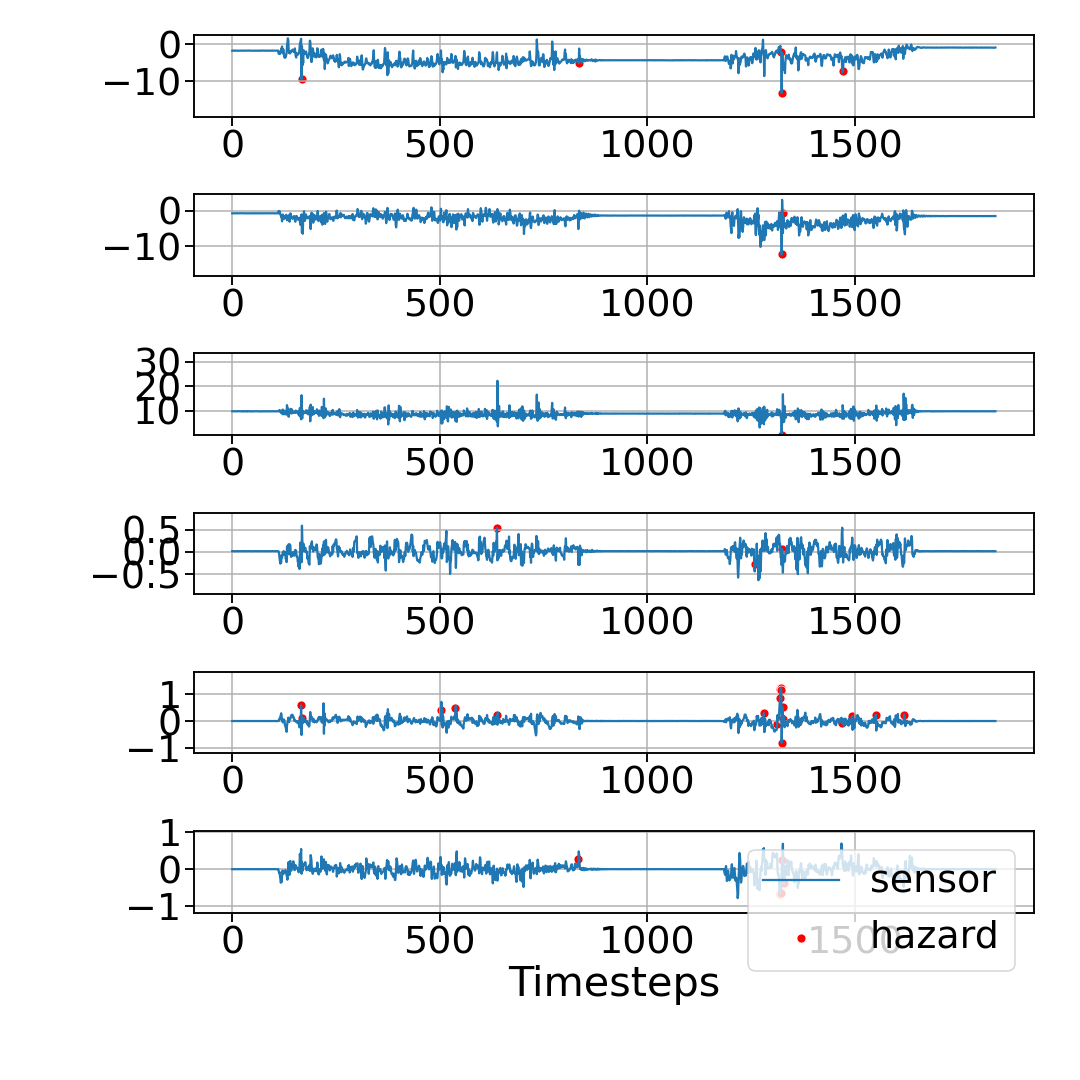}
        \caption{Test trial 3}
        \label{fig:trial2_MAE_anomaly_plot}
    \end{subfigure}
    \caption[Risk Detection with MAE]{Tip-over risk detection using MAE threshold values. IMU sequences from three test trials in blue, with red dots marking the time steps where the predictive error exceeded the detection threshold.}
    \label{fig:detection_mae}
\end{figure}

Risk event detection is triggered when the prediction error exceeds the defined thresholds. The results shown in Figures \ref{fig:detection_mse} and \ref{fig:detection_mae} highlight interesting differences between the two error metrics used. Each metric detects different nuances of anomalies: MSE penalizes errors more heavily than MAE, which becomes influential for the following reasons. The figures in \ref{fig:detection_mse} demonstrate how the MSE-based detector is more sensitive to abrupt mobility changes in the AsguardIV rover. As the rover ascends and descends a slope, the detector identifies natural motion instability while traversing the obstacle-laden slopes and craters. Additionally, it successfully detects the significant tip-over risk in test trials 1 and 3, as well as the slip that occurs in Test Trial 2. In contrast, Figure \ref{fig:detection_mae} illustrates how the MAE-based detector detects fewer minor abrupt motions but accurately detects the tip-over risks.

\section{Conclusion}

This work presented an approach for tip-over risk detection using Autoencoders for the space rover AsguardIV. The motivation of this works stemed from the need of enhanced autonomy modules for space rovers to overcome the challenges of mission monitoring. For this application, the goal was to forecast IMU signals and detect tip-over risks. The Autoencoders are trained with navigation sequences while traversing a reproduction of a lunar slope furnished with obstacles and craters.

Tip-over risk detection is consistently registered at nearly the same time steps across all sensor axes, indicating that the Autoencoder has learned to detect nuances in each sensor feature. The comparison between MSE and MAE demonstrates that both metrics are useful, depending on the degree of risk one considers more important to characterize.

The detector’s performance is intrinsically linked to the forecast performance of the Autoencoder. Therefore, selecting a robust Autoencoder was crucial. The chosen model achieved an impressive $R^{2}$ score of $0.969$ with a single-layer LSTM, fulfilling the desirable goal of high-performance requirement for deployment. This application exemplifies how tip-over risk forecast and detection can be invaluable in autonomous space robotics for navigation scenarios. 

\section*{Acknowledgments}
This work was supported by the Marie Curie Initial Training Networks (ITN) action as part of Grant FP7-608096, "ROBOCADEMY". The author was with Heriot-Watt University during the development of this work which is based on Chapter 6 of the author's PhD thesis \cite{mythesis}.

\bibliographystyle{unsrt}  
\bibliography{references}

\end{document}